\documentclass[conference]{IEEEtran}
\IEEEoverridecommandlockouts
\usepackage{algorithm}
\usepackage{algorithmic}
\usepackage{hyperref}
\hypersetup{
    colorlinks=true,
    linkcolor=blue,
    citecolor=blue,
    urlcolor=blue
}
\usepackage{cite}
\usepackage{amsmath,amssymb,amsfonts}
\usepackage{algorithmic}
\usepackage{graphicx}
\usepackage{textcomp}
\usepackage{xcolor}
\def\BibTeX{{\rm B\kern-.05em{\sc i\kern-.025em b}\kern-.08em
    T\kern-.1667em\lower.7ex\hbox{E}\kern-.125emX}}
\begin{document}

\title{Nuanced Emotion Recognition Based on a Segment-based MLLM Framework Leveraging Qwen3-Omni for A/H Detection
}

\author{\IEEEauthorblockN{1\textsuperscript{st} Liang Tang}
\IEEEauthorblockA{
\textit{Lenovo} \\
Beijing, China \\
tangliang5@lenovo.com}
\and
\IEEEauthorblockN{2\textsuperscript{nd} Hongda Li}
\IEEEauthorblockA{
\textit{Lenovo} \\
Beijing, China \\
lihd10@lenovo.com}
\and
\IEEEauthorblockN{3\textsuperscript{rd} Jiayu Zhang}
\IEEEauthorblockA{
\textit{Lenovo} \\
Beijing, China \\
zhangjy80@lenovo.com}
\and
\IEEEauthorblockN{4\textsuperscript{th} Long Chen}
\IEEEauthorblockA{
\textit{Lenovo} \\
Beijing, China \\
chenlong12@lenovo.com}
\and
\IEEEauthorblockN{5\textsuperscript{th} Shuxian Li}
\IEEEauthorblockA{
\textit{Lenovo} \\
Beijing, China \\
lisx14@lenovo.com}
\and
\IEEEauthorblockN{6\textsuperscript{th} Siqi Pei}
\IEEEauthorblockA{
\textit{Lenovo} \\
Beijing, China \\
peisq2@lenovo.com}
\and
\IEEEauthorblockN{7\textsuperscript{th} Tiaonan Duan}
\IEEEauthorblockA{
\textit{Lenovo} \\
Beijing, China \\
duantn1@lenovo.com}
\and
\IEEEauthorblockN{8\textsuperscript{th} Yuhao Cheng}
\IEEEauthorblockA{
\textit{Lenovo} \\
Beijing, China \\
yuhao.cheng@outlook.com}
}
\maketitle

\begin{abstract}
Emotion recognition in videos is a pivotal task in affective computing, where identifying subtle psychological states such as Ambivalence and Hesitancy holds significant value for behavioral intervention and digital health. Ambivalence and Hesitancy states often manifest through cross-modal inconsistencies such as discrepancies between facial expressions, vocal tones, and textual semantics, posing a substantial challenge for automated recognition. This paper proposes a recognition framework that integrates temporal segment modeling with Multimodal Large Language Models. To address computational efficiency and token constraints in long video processing, we employ a segment-based strategy, partitioning videos into short clips with a maximum duration of 5 seconds. We leverage the Qwen3-Omni-30B-A3B model, fine-tuned on the BAH dataset using LoRA and full-parameter strategies via the MS-Swift framework, enabling the model to synergistically analyze visual and auditory signals. Experimental results demonstrate that the proposed method achieves an accuracy of 85.1\% on the test set, significantly outperforming existing benchmarks and validating the superior capability of Multimodal Large Language Models in capturing complex and nuanced emotional conflicts. The code is released at \url{https://github.com/dlnn123/A-H-Detection-with-Qwen-Omni.git}.
\end{abstract}

\begin{IEEEkeywords}
Emotion recognition, Ambivalence and Hesitancy, Multimodal Large Language Models, Qwen3-Omni, Video processing.
\end{IEEEkeywords}

\section{Introduction}

Emotion recognition\cite{Facial-Expression} in videos has become an important research topic in affective computing, with applications in digital health\cite{Emotion-Recognition}, human--computer interaction, and behavioral analysis. In particular, recognizing subtle psychological states such as Ambivalence and Hesitancy (A/H)\cite{gonzalez-gonzalez2026bah} is critical in behavioral change interventions. A/H reflects a conflicting emotional state in which individuals simultaneously experience positive and negative\cite{Beyond-attitudinal} attitudes toward a behavior or decision, often leading to delayed or abandoned behavioral change. Detecting such states automatically can support digital health systems, enabling adaptive interventions and scalable behavior change programs.

Despite its importance, automatic recognition of A/H remains challenging\cite{ryumina2025teamras9thabaw}. Unlike basic emotions such as happiness or sadness, A/H manifests as a subtle and conflicting affective state, often expressed through inconsistencies across modalities\cite{Guided-Interpretable,liu2024norfaceimprovingfacialexpression,Xue_2023}, including facial expressions, speech patterns, language cues, and body movements. These multimodal conflicts can occur both across modalities (e.g., positive speech with hesitant tone) and within a modality (e.g., fluctuating facial expressions), making the task particularly difficult for conventional emotion recognition systems.

The recently introduced BAH dataset\cite{gonzalez-gonzalez2026bah} provides a benchmark for studying A/H recognition in videos. It contains 1,427 videos from 300 participants, totaling approximately 10.6 hours of recordings, with expert annotations at both video and frame levels. Participants respond to predefined behavioral questions designed to elicit ambivalent or hesitant responses, producing naturalistic multimodal signals such as facial expressions, audio cues, and spoken language transcripts. However, existing baseline approaches demonstrate limited performance, highlighting the difficulty of the task and the need for more effective modeling strategies.

A key challenge in applying modern multimodal large models to this task lies in the temporal length of video inputs. Directly feeding long videos into large multimodal models often exceeds token limits and increases computational cost, while also introducing irrelevant contextual noise. Meanwhile, previous studies indicate that A/H signals are typically short temporal segments, with an average duration of approximately 4.29 seconds within videos. This observation suggests that modeling shorter temporal windows may be sufficient for detecting A/H while improving computational efficiency.
\begin{figure*}[t]
  \centering  
  \includegraphics[width=\textwidth]{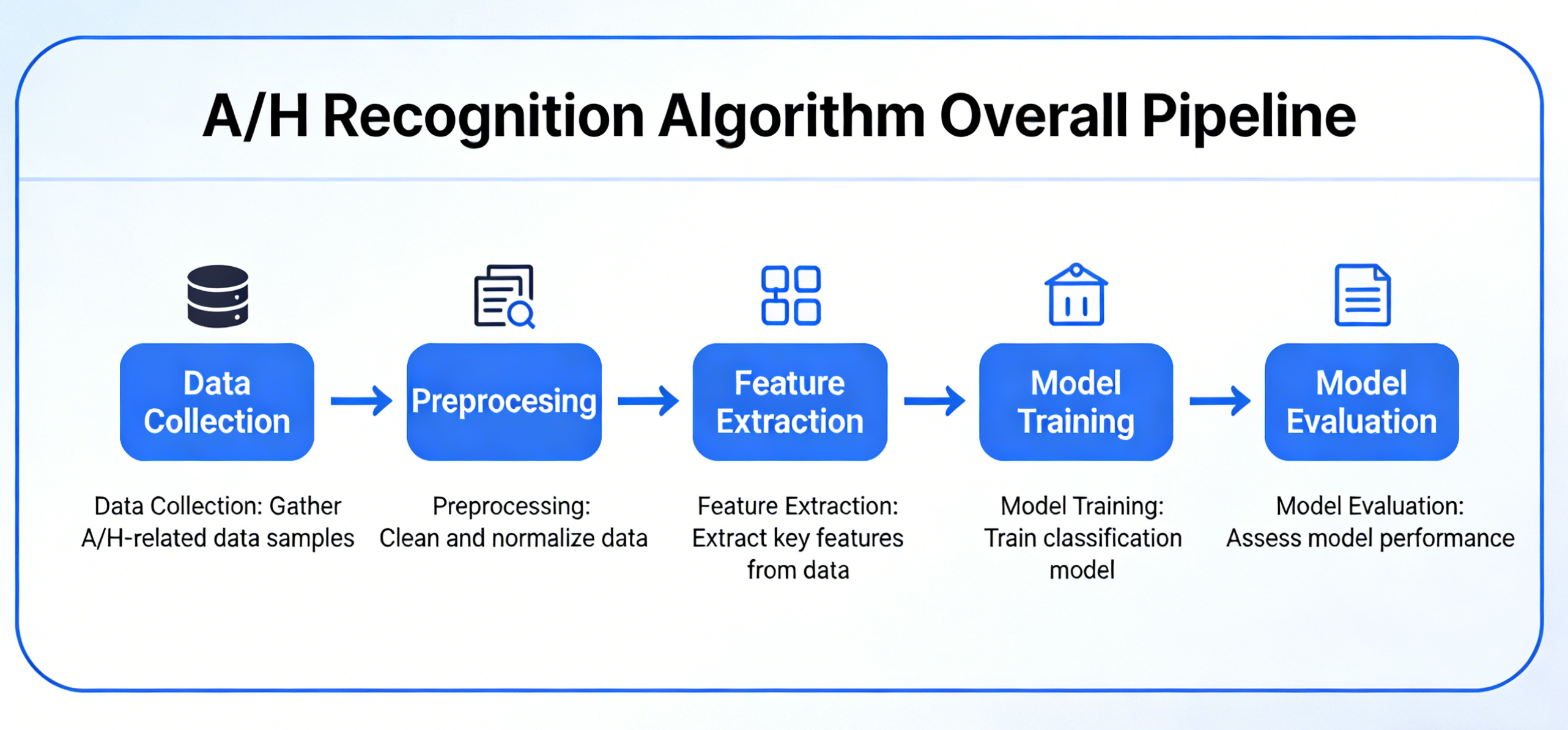}  
  \caption{A/H Recognition Algorithm Overall Pipeline}
  \label{fig:algorithm}
\end{figure*}

To model multimodal interactions effectively, we adopt Qwen3-Omni-30B-A3B\cite{Qwen3-Omni}, a powerful multimodal large language model capable of integrating visual and audio information. The model is fine-tuned on the BAH dataset using the proposed clip-based strategy, enabling it to capture subtle cross-modal cues and temporal inconsistencies associated with ambivalence and hesitancy. Experimental results demonstrate that our approach significantly improves recognition performance, achieving 85.1\% accuracy on the test set, substantially outperforming previously reported baselines.

In summary, our work makes the following contributions:
\begin{itemize}
    \item \textbf{Clip-based temporal modeling:} We propose a simple yet effective strategy that segments videos into 5-second clips, allowing multimodal models to capture local A/H signals while respecting token constraints.
    
    \item \textbf{Multimodal large model adaptation:} We adapt Qwen3-Omni-30B-A3B for the A/H recognition task, leveraging its multimodal reasoning capability to detect subtle cross-modal emotional conflicts.
    
    \item \textbf{Strong empirical performance:} Our method achieves 85.1\% accuracy on the test set, demonstrating the effectiveness of combining temporal segmentation with multimodal large models for A/H recognition.
\end{itemize}

These results suggest that large multimodal models, when combined with appropriate temporal segmentation strategies, can significantly advance the detection of complex and subtle emotional states such as ambivalence and hesitancy.

\section{Method}

In this section, we describe the proposed framework for Ambivalence/Hesitancy (A/H) recognition. As shown in Fig.1, the overall pipeline consists of three stages: (1) A/H-aware video preprocessing based on annotation timestamps, (2) clip-based temporal segmentation, and (3) multimodal large model fine-tuning and inference using Qwen3-Omni-30B-A3B.

\subsection{A/H-Aware Video Preprocessing}

The BAH dataset provides detailed annotations indicating whether a video contains A/H and, if present, the temporal segments where A/H occurs. We leverage this information to construct more informative training samples.

Specifically, for each video we first read the annotation file that includes a global A/H label and detailed time intervals describing where A/H appears. If the video does not contain A/H (i.e., the global label is negative), the entire video is retained as a single training sample. In contrast, if the video contains A/H, we extract only the annotated temporal segments corresponding to A/H occurrences.

Formally, for a video $V$ with duration $T$, the annotation file provides a set of temporal segments:
\begin{equation}
S = \{(t_1^\text{start}, t_1^\text{end}), (t_2^\text{start}, t_2^\text{end}), \ldots, (t_k^\text{start}, t_k^\text{end})\}.
\end{equation}

Each segment is cropped using \texttt{ffmpeg} to produce a new video clip that focuses on the time interval where A/H signals appear. This preprocessing step significantly reduces irrelevant visual and acoustic information while increasing the density of meaningful A/H cues in the training data.

\subsection{Clip-Based Temporal Segmentation}

Although A/H segments are shorter than full videos, their duration can still vary significantly. Directly feeding long videos into a multimodal large model is inefficient and may exceed the model's token limits. To address this issue, we adopt a clip-based temporal segmentation strategy.

Given a video segment with duration $T$, we divide it into multiple shorter clips with a fixed maximum length of 5 seconds. The segmentation process follows a uniform sliding strategy:

\begin{itemize}
\item If $T < 5s$, the clip is kept unchanged.
\item If $T \geq 5s$, the video is divided into consecutive clips of 5 seconds.
\item Very short trailing segments are discarded to avoid extremely short clips.
\end{itemize}

This strategy ensures that each input sample contains a compact temporal window that is sufficient to capture subtle emotional cues while remaining within the token capacity of the multimodal model.

\subsection{Multimodal Large Model Fine-Tuning}

To model cross-modal emotional signals, we adopt \textbf{Qwen3-Omni-30B-A3B}, a large multimodal language model capable of jointly processing visual and audio inputs. The model integrates video frames, audio features, and textual prompts within a unified transformer architecture.

\begin{algorithm}[t]
\caption{Overall Pipeline for A/H Recognition}
\label{alg:overall_pipeline}
\begin{algorithmic}[1]
\REQUIRE Raw video set $\mathcal{V}$, annotation file $\mathcal{A}$, clip length $\Delta=5$s, multimodal model $M$
\ENSURE Video-level A/H predictions

\STATE Initialize processed dataset $\mathcal{D} \leftarrow \emptyset$

\FOR{each video $V_i \in \mathcal{V}$}
    \STATE Read annotation info $a_i$ from $\mathcal{A}$
    
    \IF{$a_i$ indicates A/H-negative}
        \STATE Keep the whole video as one sample
        \STATE $\hat{V}_i \leftarrow V_i$
        \STATE Split $\hat{V}_i$ into clips of length at most $\Delta$
        \STATE Assign label $y_i = 0$ to all clips
        \STATE Add all clips into $\mathcal{D}$
    \ELSE
        \STATE Obtain annotated A/H time intervals $\{(t_k^{start}, t_k^{end})\}_{k=1}^{K}$
        \FOR{each annotated interval $(t_k^{start}, t_k^{end})$}
            \STATE Crop sub-video $\hat{V}_{ik}$ from $V_i$ using $(t_k^{start}, t_k^{end})$
            \STATE Split $\hat{V}_{ik}$ into clips of length at most $\Delta$
            \STATE Assign label $y_i = 1$ to all clips
            \STATE Add all clips into $\mathcal{D}$
        \ENDFOR
    \ENDIF
\ENDFOR

\STATE Construct multimodal instruction-tuning samples using video, audio, and task prompt
\STATE Fine-tune model $M$ on $\mathcal{D}$

\FOR{each test video $V_j$}
    \STATE Split $V_j$ into clips $\{c_1, c_2, \dots, c_N\}$
    \FOR{each clip $c_n$}
        \STATE Build multimodal input with video, audio, and prompt
        \STATE Use $M$ to predict clip label $\hat{y}_n \in \{0,1\}$
    \ENDFOR
    \STATE Aggregate clip predictions:
    \[
    \hat{y}_{video} = \max(\hat{y}_1, \hat{y}_2, \dots, \hat{y}_N)
    \]
\ENDFOR

\RETURN Video-level predictions for all test videos
\end{algorithmic}
\end{algorithm}
We fine-tune the model using the \textbf{MS-Swift} training framework with both parameter-efficient tuning (LoRA) and full fine-tuning strategies. During training, each clip is converted into a multimodal conversation format consisting of video input, audio input, and a task-specific prompt.

The model is trained to answer a binary question, and the expected output is restricted to the format:

\begin{center}
\texttt{<answer>Yes</answer>} or \texttt{<answer>No</answer>}
\end{center}

This formulation transforms A/H recognition into a multimodal reasoning task where the model jointly analyzes visual expressions, speech characteristics, and temporal context.




\subsection{Inference and Video-Level Prediction}

During inference, each video is processed using the same clip segmentation strategy. For every clip, the model generates a binary prediction indicating whether A/H is present.

Given a video consisting of $N$ clips, the final video-level prediction is obtained through an aggregation rule. If any clip is predicted as containing A/H, the entire video is classified as positive:

\begin{equation}
y_{\text{video}} = \max_{i=1}^{N} y_i
\end{equation}

where $y_i$ represents the prediction for the $i$-th clip.

This strategy enables the model to capture localized A/H signals while producing stable video-level predictions.

\paragraph{Multi-model Voting Strategy.}
To further improve prediction robustness, we adopt a multi-model ensemble strategy during inference. Specifically, multiple models are trained using different fine-tuning configurations, including LoRA-based tuning, full-parameter fine-tuning, and variations in training hyperparameters. 

During testing, each model independently produces a video-level prediction following the clip aggregation procedure described above. The final prediction is then determined by a majority voting rule across all models. Let $\hat{y}^{(k)}$ denote the prediction from the $k$-th model. The final decision is obtained as

\begin{equation}
\hat{y} = \mathrm{MajorityVote}(\hat{y}^{(1)}, \hat{y}^{(2)}, \dots, \hat{y}^{(K)})
\end{equation}

where $K$ is the number of models in the ensemble.

This ensemble strategy reduces prediction variance and improves the stability of the final decision, especially for ambiguous samples where individual models may produce inconsistent outputs.

\section{Experiment}

In this section, we evaluate the effectiveness of the proposed method on the BAH dataset. We first introduce the experimental setup and implementation details, followed by comparisons with different model configurations and training strategies. Finally, we analyze the results and discuss key observations.

\subsection{Experimental Setup}

All experiments are implemented using the \textbf{MS-Swift} training framework. The backbone model used in our method is \textbf{Qwen3-Omni-30B-A3B}, a multimodal large language model capable of jointly processing video and audio inputs.

The training dataset is constructed using the preprocessing strategy described in Section~3. Specifically, videos containing A/H annotations are cropped according to the provided timestamps, while videos without A/H annotations are kept unchanged. All resulting videos are further segmented into clips with a maximum duration of 5 seconds.

Each clip is converted into a multimodal instruction-following sample consisting of:

\begin{itemize}
\item Video input
\item Audio input
\item Task prompt asking whether the clip contains A/H emotion
\end{itemize}

During inference, predictions are first obtained at the clip level and then aggregated into a final video-level prediction using a max-based aggregation rule.

\subsection{Implementation Details}

All experiments are implemented using the \textbf{MS-Swift} framework and conducted on multiple GPUs with mixed-precision training using \texttt{bfloat16}. To improve efficiency for long-context multimodal inputs, we enable \texttt{FlashAttention}. The maximum sequence length is set to 32768 tokens.

We evaluate two fine-tuning strategies: \textbf{LoRA fine-tuning} and \textbf{full fine-tuning}.

For \textbf{LoRA fine-tuning}, the learning rate is set to $1\times10^{-5}$ and the model is trained for 1 epoch. The LoRA adapters are configured with rank 8 and alpha 32. The per-device batch size is 2, and the gradient accumulation steps are set to 32.

For \textbf{full fine-tuning}, all model parameters are updated. To ensure stable optimization, we use a smaller learning rate of $1\times10^{-6}$ and train the model for 2--3 epochs. The per-device batch size and gradient accumulation steps remain the same as in the LoRA setting.

In addition, we explore two different \textbf{prompt formulations} during full fine-tuning. Both prompts instruct the model to determine whether the input video clip exhibits Ambivalence/Hesitancy emotion, but they differ in wording and instruction structure. This allows us to evaluate the impact of prompt design on multimodal reasoning performance.

The performance of all models is evaluated using Accuracy, Precision, Recall, and F1-score.

\begin{table}[h]
\centering
\caption{Training hyperparameters for different fine-tuning strategies.}
\label{tab:train_hyper}
\begin{tabular}{lcc}
\hline
Hyperparameter & LoRA & Full Fine-tuning \\
\hline
Learning rate & $1\times10^{-5}$ & $1\times10^{-6}$ \\
Epochs & 1 & 2--3 \\
Per-device batch size & 2 & 2 \\
Gradient accumulation & 32 & 32 \\
LoRA rank & 8 & -- \\
LoRA alpha & 32 & -- \\
Precision & bfloat16 & bfloat16 \\
FlashAttention & Yes & Yes \\
Max length & 32768 & 32768 \\
\hline
\end{tabular}
\end{table}

\subsection{Model Comparison}

To evaluate the effectiveness of the proposed pipeline, we conduct experiments using multiple model configurations and training strategies. These include different fine-tuning methods and variations of training parameters.

Table~\ref{tab:model_comparison} summarizes the accuracy obtained by different models.

\begin{table}[h]
\centering
\caption{Performance comparison of different model configurations on the test set.}
\label{tab:model_comparison}
\begin{tabular}{l c}
\hline
Model Configuration & Accuracy \\
\hline
Qwen3-Omni-v1 & 81.9\% \\
Qwen3-Omni-v2 & 79.8\% \\
Qwen3-Omni-v3 & 65.3\% \\
Qwen3-Omni (Majority Vote) & 85.1\% \\
\hline
\end{tabular}
\end{table}

In this comparison, \textbf{Qwen3-Omni-v1} corresponds to the LoRA fine-tuning setting trained for one epoch. 
\textbf{Qwen3-Omni-v2} and \textbf{Qwen3-Omni-v3} are obtained through full-parameter fine-tuning with different training epochs, where v2 is trained for 2 epochs and v3 for 3 epochs.

To further improve prediction robustness, we combine these models using a \textbf{majority voting strategy}. 
Specifically, each model independently produces a video-level prediction, and the final decision is determined by the majority of model outputs. 
This ensemble approach helps reduce prediction variance and improves stability, especially for ambiguous samples.

The results show that combining multiple fine-tuned models through voting further improves the overall performance compared with individual models.

\section{Conclusion}
In this study, we proposed a robust deep learning framework tailored for the recognition of complex Ambivalence/Hesitancy (A/H) emotions in video data. By implementing a temporal segmentation strategy that divides videos into short segments of $T \le 5s$, the approach effectively overcomes the computational bottlenecks associated with long video inputs while focusing on local windows where A/H signals are most concentrated. The successful adaptation of the Qwen3-Omni-30B-A3B model facilitates the capture of subtle discrepancies between facial expressions and speech through its advanced cross-modal reasoning capabilities. Furthermore, the integration of a max-based aggregation rule, defined as $y_{video} = \max_{i=1}^N y_i$, alongside a multi-model majority voting strategy, significantly enhances the system's robustness when processing ambiguous samples. The experimental results confirm that our method achieves superior accuracy and stability, providing a solid technical foundation for integrating sophisticated emotion recognition into future digital health and behavioral analysis systems.
\bibliographystyle{IEEEtran}
\bibliography{IEEE-refrence}
\end{document}